\documentclass[11pt,a4paper]{article}
\usepackage{authblk}

\usepackage{naaclhlt2019}

\usepackage{times}
\usepackage{latexsym}
\usepackage{amsmath} 
\usepackage{amssymb}
\usepackage{graphicx}
\usepackage{subcaption}
\usepackage{booktabs}
\usepackage{pifont}
\usepackage{xcolor}
\usepackage{booktabs}
\usepackage{tabularx}

\usepackage{url}

\aclfinalcopy 
\def\aclpaperid{6} 

\title{Composition of Sentence Embeddings:\\Lessons from Statistical Relational Learning}
\author[3,1]{Damien Sileo}
\author[2]{Tim Van De Cruys}
\author[1]{Camille Pradel}
\author[3]{Philippe Muller}

\affil[1]{Synapse D\'eveloppement, Toulouse, France}
\affil[2]{IRIT, CNRS, France} \vspace{-2ex}
\affil[3]{IRIT, University of Toulouse, France}
\affil[ ]{\url{damien.sileo@synapse-fr.com}}

\date{}

\begin{document}
\maketitle
\begin{abstract}

Various NLP problems -- such as the prediction of sentence similarity, entailment, and discourse relations -- are all instances of the same general task: the modeling of semantic relations between a pair of textual elements.
A popular model for such problems is to embed sentences into fixed size vectors, and use composition functions (e.g. concatenation or sum) of those vectors as features for the prediction. 
At the same time, composition of embeddings has been a main focus within the field of Statistical Relational Learning (SRL) whose goal is to predict relations between entities (typically from knowledge base triples).
In this article, we show that previous work on relation prediction between texts implicitly uses compositions from baseline SRL models. We show that such compositions are not expressive enough for several tasks (e.g. natural language inference). 
We build on recent SRL models to address textual relational problems, showing that they are more expressive, and can alleviate issues from simpler compositions. The resulting models significantly improve the state of the art in both transferable sentence representation learning and relation prediction.
\end{abstract}

\section{Introduction} 
\label{sec:intro}

Predicting relations between textual units is a widespread task, essential for discourse analysis, dialog systems, information retrieval, or paraphrase detection. Since relation prediction often requires a form of understanding, it can also be used as a proxy to learn transferable sentence representations. Several tasks that are useful to build sentence representations are derived directly from text structure,
without human annotation: 
sentence order prediction \citep{Logeswaran2016, Jernite2017}, the prediction of previous and subsequent sentences \citep{Kiros2015,Jernite2017}, or the prediction of explicit discourse markers between sentence pairs \citep{Nie2017, Jernite2017}.
Human labeled relations between sentences can also be used for that purpose, e.g. inferential relations \citep{Conneau2017}.
While most work on sentence similarity estimation, entailment detection, answer selection, or discourse relation prediction seemingly uses task-specific models, they all involve predicting whether a relation $R$ holds between two sentences $s_1$ and  $s_2$. This genericity has been noticed in the literature before \citep{Baudi} and it has been leveraged for the evaluation of sentence embeddings within the SentEval framework \citep{Conneau2017}.

A straightforward way to predict the probability of $(s_1, R, s_2) $ being true is
to represent $s_1$ and $s_2$ with $d$-dimensional embeddings $h_1$ and $h_2$, and to compute sentence pair features $f(h_1, h_2)$, where $f$ is a composition function (e.g. concatenation, product, \ldots).  A softmax classifier $g_\theta$ can learn to predict $R$ with those  features.  $g_\theta \circ f$ can be seen as a reasoning based on the content of $h_1$ and $h_2$ \citep{Socher2013}. 

Our contributions are as follows:
\begin{itemize}
\itemsep0em 

\item[--] we review composition functions used in textual relational learning and show that they lack expressiveness (section \ref{sec:comp});
\item[--] we draw analogies with existing SRL models (section \ref{sec:srl}) and design new compositions inspired from SRL (section \ref{sec:trm});
\item[--] we perform extensive experiments to test composition functions 
 and show that some of them can improve the learning of representations and their downstream uses 
(section \ref{sec:exp}).
\end{itemize}

\section{Composition functions for relation prediction}
\label{sec:comp}
We review here popular composition functions used for relation prediction based on sentence embeddings. 
Ideally, they should simultaneously fulfill  the following minimal requirements:
\begin{itemize}
\itemsep0em 
\item[--] make use of interactions between representations of sentences to relate; 
\item[--] allow for the learning of asymmetric relations (e.g. entailment, order);
\item[--] be usable with high dimensionalities (parameters $\theta$ and $f$ should fit in GPU memory).
\end{itemize}
Additionally, if the main goal is transferable sentence representation learning, compositions should also incentivize gradually changing sentences to lie on a linear manifold, since transfer usually uses linear models.
Another goal can be learning of transferable {\it relation} representation. Concretely, a sentence encoder and $f$ can be trained on a base task, and $f(h_1,h_2)$ can be used as features for transfer in another task.  In that case, the geometry of the sentence embedding space is less relevant, as long as the $f(h_1,h_2)$ space works well for transfer learning. Our evaluation will cover both cases.

A straightforward instantiation of $f$ is concatenation \citep{Hooda2017}:
\begin{equation}
f_{[,]}(h_1,h_2)= [h_1, h_2]
\end{equation}
However, {\it interactions} between $s_1$ and $s_2$ cannot be modeled with $f_{[,]}$ followed by a softmax regression. Indeed, $f_{[,]}(h_1,h_2)\theta$  can be rewritten as a sum of independent contributions from $h_1$ and $h_2$, namely $\theta_{[0:d]} h_1 +\theta_{[d:2d]} h_2$.
Using a multi-layer perceptron  before the softmax would solve this issue, but it also harms sentence representation learning \citep{Conneau2017,Logeswaran2018}, possibly because the perceptron allows for accurate predictions even if the sentence embeddings lie in a convoluted space.  
%
%
To promote interactions between $h_1$ and $h_2$,  element-wise product has been used in \citet{Baudi}:
\begin{equation}
\label{eq:hadcomposition}
f_{\odot}(h_1,h_2)= h_1 \odot h_2 
\end{equation}
Absolute difference is another solution for sentence similarity \citep{mueller2016siamese}, and its element-wise variation may equally be used to compute informative features:
\begin{equation}
\label{eq:abs}
f_{-}(h_1,h_2)= |h_1 - h_2| 
\end{equation}
The latter two were combined into a popular instantiation, sometimes refered as \textit{heuristic matching} \citep{Tai2015,Kiros2015,Mou2015}:
\begin{equation}
\label{eq:vanillacomposition}
f_{\odot-}(h_1,h_2)= [h_1 \odot h_2, |h_2-h_1|]
\end{equation} Although effective for certain similarity tasks, $f_{\odot-}$ is symmetrical, and should be a poor choice for tasks like entailment prediction or prediction of discourse relations. For instance, if $R_e$ denotes entailment and $(s_1, s_2)$= (``It just rained", ``The ground is wet"), $(s_1, R_e, s_2)$ should hold but not $(s_2, R_e, s_1)$. The $f_{\odot -}$ composition function is nonetheless 
used to train/evaluate models on entailment \citep{Conneau2017} or discourse relation prediction \citep{Nie2017}.

Sometimes $[h_1, h_2]$ is concatenated to $f_{\odot-}(h_1,h_2)$ \citep{Ampomah2016, Conneau2017}. While the resulting composition is asymmetrical, the asymmetrical component involves no interaction as noted previously. 
We note that this composition is very commonly used. On the SNLI benchmark,\footnote{\url{nlp.stanford.edu/projects/snli/}, as of February 2019.} $12$ out of the $25$ listed sentence embedding based models use it, and $7$ use a weaker form (e.g. omitting $f_{\odot}$).

The outer product $\otimes$ has been used instead for asymmetric multiplicative interaction \citep{Jernite2017}:
\begin{equation}
f_{\otimes}(h_1, h_2) = h_1 \otimes  h_2 \text{ where } (h_1 \otimes  h_2)_{i,j} = h_{1i} h_{2j}
\end{equation}
 This formulation is expressive but it forces $g_{\theta}$ to have $d^2$ parameters per relation, which is prohibitive when there are many relations and $d$ is high. 

The problems outlined above are well known in SRL. Thus, existing compositions (except $f_{\otimes}$) can only model relations superficially for tasks currently used to train state of the art sentence encoders, like NLI or discourse connectives prediction.

\section{Statistical Relational Learning models}
\label{sec:srl}
 \begin{table}[t]
\centering
\begin{footnotesize}
\begin{tabular}{lllll}
\toprule
Model    & Scoring function                       & Parameters       &  &  \\
\midrule
Unstructured  &  $||e_1-e_2||_p$&- & &\\
TransE  & $||e_1 + w_r - e_2||_p$                  & $w_r \in \mathbb R^d$     &  &  \\
RESCAL   & $e_1^T  W_r e_2$                       & $W_r \in \mathbb R^{d^2}$ &  &  \\
DistMult & $ <e_1, w_r, e_2>$                     & $w_r \in \mathbb R^d$     &  &  \\
ComplEx   & $ \text{Re}<e_1, w_r, \overline{e_2}>$ & $w_r \in \mathbb C^d$     &  & \\
\bottomrule
\end{tabular}
\caption{Selected relational learning models. 
Unstructured is from \citep{Bordes2013a}, TransE from \citep{Bordes2013}, RESCAL from \citep{Nickel2011}, DistMult from \citep{Yang2015} and \citep{Trouillon2016}.
Following the latter, $< a, b, c >$ denotes $ \sum_k a_k b_k c_k
  . $ $\text{Re}(x)$  is the real part of $x$, and $p$ is commonly set to $1$.} \label{tab:relmodels}
  \end{footnotesize}
\end{table}
In this section we introduce the context of statistical relational learning (SRL) and relevant models. Recently, SRL has focused on efficient and expressive relation prediction based on embeddings. 
A core goal of SRL \citep{Getoor:2007:ISR:1296231} is to induce whether a relation $R$ holds between two arbitrary entities $e_1, e_2$.
As an example, we would like to assign a score to $(e_1, R, e_2)$ = (Paris, {\sc located\_in}, France) that reflects a high probability. 
In embedding-based SRL models, entities $e_i$ have vector representations in $\mathbb{R}^d$ and a scoring function reflects truth values of relations. The scoring function should allow for relation-dependent reasoning over the latent space of entities. Scoring functions can have relation-specific parameters, which can be interpreted as relation embeddings. 
Table \ref{tab:relmodels} presents an overview of a number of state of the art relational models. 
We can distinguish two families of models: subtractive and multiplicative.

The TransE scoring function is motivated by the idea that translations in latent space can model  analogical reasoning and hierarchical relationships. Dense word embeddings trained on tasks related to the distributional hypothesis naturally allow for analogical reasoning with translations without explicit supervision \citep{Mikolov2013}. TransE generalizes the older Unstructured model. We call them subtractive models.
%
%

The RESCAL, Distmult, and ComplEx scoring functions can be seen as dot product matching between $e_1$ and a relation-specific linear transformation of $e_2$ \citep{Liu2017a}. This transformation helps checking whether $e_1$ matches with some aspects of $e_2$. 
RESCAL allows a full linear mapping $W_r e_2 $ but has a high complexity, while Distmult is restricted to a component-wise weighting $w_r \odot e_2$.
ComplEx has fewer parameters than RESCAL but still allows for the modeling of asymmetrical relations.  
As shown in \citet{Liu2017a}, ComplEx boils down to a restriction of RESCAL where $W_r$ is a block diagonal matrix.  These blocks are 2-dimensional, antisymmetric and have equal diagonal terms. 
%
Using such a form, even and odd indexes of $e$'s dimensions play the roles of real and imaginary numbers respectively.
The ComplEx model \citep{Trouillon2016}  and its variations \citep{Lacroix2018CanonicalTD} yield state of the art performance on knowledge base completion on numerous evaluations.

\section{Embeddings composition as SRL models}
\label{sec:trm}
\begin{figure}[]
  \centering
\includegraphics[width =0.45\textwidth]{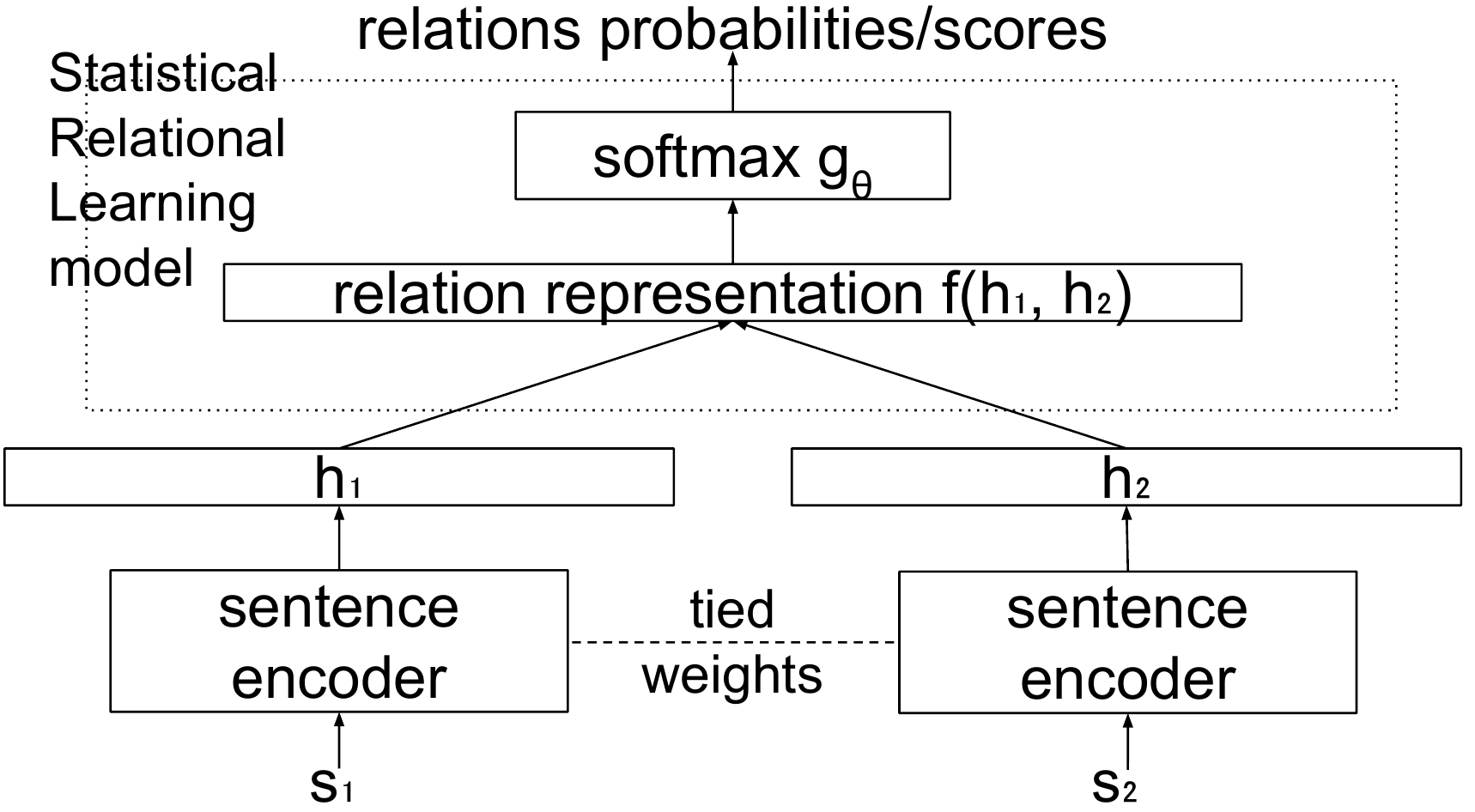}
  \caption{Implicit SRL model in text relation prediction}
\label{ov} 
\end{figure}

We claim that several existing models \citep{Conneau2017,Nie2017,Baudi} boil down to SRL models where the sentence embeddings ($h_1, h_2)$ act as entity embeddings ($e_1,e_2$). This framework is depicted in figure \ref{ov}.
In this article we focus on sentence embeddings, although our framework can straightforwardly be applied to other levels of language granularity (such as words, clauses, or documents). 

Some models \citep{conf/acl/ChenZLWJI17,DBLP:journals/corr/SeoKFH16,gong2018natural,Radford2018ImprovingLU,devlin2018bert} do not rely on explicit sentence encodings to perform relation prediction. They combine information of input sentences at earlier stages, using conditional encoding or cross-attention. There is however no straightforward way to derive transferable sentence representations in this setting, and so these models are out of the scope of this paper. They sometimes make use of composition functions, so our work could still be relevant to them in some respect. 

In this section we will make a link between sentence composition functions and SRL scoring functions, and propose new scoring functions drawing inspiration from SRL. 
\begin{figure*}[htb!]
  
  \begin{subfigure}[t]{0.49\textwidth}
    \begin{center}  
      \includegraphics[scale=0.5]{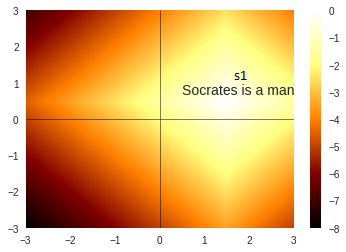}
      \caption{Score map of $(s_1, R_{to\_the\_past}, s_2$) over possible sentences $s_2$ using Unstructured composition.}
      \label{fig1a}
    \end{center}
  \end{subfigure}
  \begin{subfigure}[t]{0.49\textwidth}
        \begin{center}  
\includegraphics[scale=0.5]{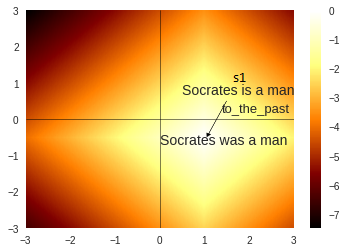}  
    \subcaption{Score map of $(s_1, R_{to\_the\_past}, s_2$) over possible sentences $s_2$ using TransE composition.}
          \label{fig1b}
    \end{center}
    \end{subfigure}
  \begin{subfigure}[b]{0.49\textwidth}
    \begin{center}  
    \includegraphics[scale=0.5]{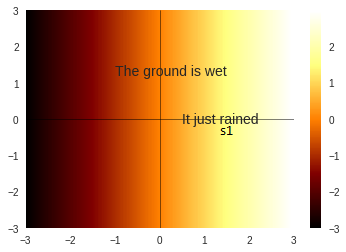}
  \subcaption{Score map of $(s_1, R_{entailment}, s_2$) over possible sentences $s_2$ using DistMult composition.}
        \label{fig2a}
    \end{center}
  \end{subfigure}
  \begin{subfigure}[b]{0.49\textwidth}
    \begin{center}  
    \includegraphics[scale=0.5]{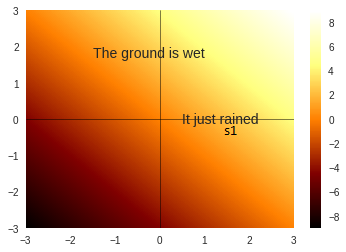}  
    \subcaption{Score map of $(s_1, R_{entailment}, s_2$) over possible sentences $s_2$ using ComplEx composition.}
          \label{fig2b}
    \end{center}
    \end{subfigure}
    \caption{Possible scoring function values according to different composition functions. $s_1$ and $R$ are fixed and color brightness reflects likelihood of $(s_1, R, s_2)$ for each position of embedding $s_2$. (b) and (d) are respectively more expressive than (a) and (c).}\label{fg:comparison}
\end{figure*}
\subsection{Linking composition functions and SRL models}
 The composition function $f_{\odot}$ from equation \ref{eq:hadcomposition} followed by a softmax regression yields a score whose analytical form is identical to the Distmult model score described in section \ref{sec:srl}.
Let $\theta_R$ denote the softmax weights for relation $R$. The logit score for the truth of $(s_1, R, s_2)$ is
 $ f(h_1,h_2)\theta_R = (h_1 \odot h_2)\theta_R$ which is equal to the Distmult scoring function $<h_1, \theta_R, h_2>$ if $h_1,h_2$ act as entities embeddings and $\theta_R$ as the relation weight $w_R$ . 
 
Similarly, the composition  $f_{-}$ from equation \ref{eq:abs} followed by a softmax regression can be seen as an element-wise weighted score of Unstructured (both are equal if softmax weights are all unitary).

Thus, $f_{\odot -}$ from \ref{eq:vanillacomposition} (with softmax regression) can  be seen as a weighted ensemble of Unstructured and Distmult. These two models are respectively outperformed by TransE and  ComplEx on knowledge base link prediction by a large margin \citep{Trouillon2016,Bordes2013a}. We therefore propose to change the Unstructured and Distmult in $f_{\odot -}$ such that they match their respective state of the art variations in the following sections. We will also show the implications of these refinements.
\subsection{Casting TransE as a composition}
Simply replacing $|h_2-h_1|$ with
\begin{equation}
f_t(h_1,h_2) = |h_2-h_1 +t |  \textrm{, where } t \in \mathbb R^d 
\end{equation} would make the model analogous to TransE. $t$ is learned and is shared by all relations.
A relation-specific translation $t_R$ could be used but it would make $f$ relation-specific. 
Instead, here, each dimension of $f_t(h_1,h_2) $ can be weighted according to a given relation.
Non-zero $t$ makes $f_t$ asymmetrical and also yields features that allow for the checking of an analogy between $s_1$ and $s_2$. Sentence embeddings often rely on pre-trained word embeddings which have demonstrated strong capabilities for analogical reasoning.
Some analogies, such as \textit{part-whole}, are computable with off-the-shelf word embeddings \citep{Chen2017a} and should be very informative for natural language inference tasks.
As an illustration, let us consider an artificial semantic space (depicted in figures \ref{fig1a} and \ref{fig1b}) where we posit that there is a ``to the past" translation $t$ so that $h_1 + t $ is the embedding of a sentence $s_1$ changed to the past tense. Unstructured is not able to leverage this semantic space to correctly score $(s_1, R_{to\_the\_past}, s_2$) while TransE is well tailored to provide highest scores for sentences near $h_1 + \hat{t}$ where $\hat t $ is an estimation of $t$ that could be learned from examples.

\subsection{Casting ComplEx as a composition}
Let us partition $h$ dimensions into two equally sized sets $\mathcal{R}$ and $\mathcal{I}$, e.g. even and odd dimension indices of $h$.
We propose a new function $f_\mathbb{C}$ as a way to fit the ComplEx scoring function into a composition function.
\begin{equation}
\begin{split}
 f_\mathbb{C}(h_1, h_2) = [
h_1^\mathcal{R} \odot h_2^\mathcal{R} +   h_1^\mathcal{I} \odot  h_2^\mathcal{I}  ,\\
h_1^\mathcal{R}  \odot h_2^\mathcal{I} -  h_1^\mathcal{I} \odot h_2^\mathcal{R}
]
\end{split}
\end{equation} 
$f_\mathbb{C}(h_1, h_2)$ multiplied by softmax weights $\theta_r$ is equivalent to the ComplEx scoring function \mbox{$\text{Re}<h_1,\theta_r,\overline{h_2}>$}. The first half of $\theta_r$ weights corresponds to the real part of ComplEx relation weights  while the last half corresponds to the imaginary part.

$f_\mathbb{C}$ is to the ComplEx scoring function what $f_\odot$ is to the DistMult scoring function.
Intuitively, ComplEx is a minimal way to model interactions between distinct latent dimensions while Distmult only allows for identical dimensions to interact.

Let us consider a new artificial semantic space (shown in figures \ref{fig2a} and \ref{fig2b}) where the first dimension is high when a sentence means that it just rained, and the second dimension is high when the ground is wet.
Over this semantic space, Distmult is only able to detect entailment for paraphrases whereas ComplEx is also able to naturally model that (``it just rained", $R_{entailment}$, ``the ground is wet") should be high while its converse should not. 

We also propose two more general versions of $f_\mathbb{C}$ :

\begin{equation}
\begin{split}
 f_\mathbb{C^{\alpha}}(h_1, h_2) = [
h_1^\mathcal{R} \odot h_2^\mathcal{R} ,   h_1^\mathcal{I} \odot  h_2^\mathcal{I}  ,\\
h_1^\mathcal{R}  \odot h_2^\mathcal{I} -  h_1^\mathcal{I} \odot h_2^\mathcal{R}
]
\end{split}
\end{equation} 

\begin{equation}
\begin{split}
 f_\mathbb{C^{\beta}}(h_1, h_2) = [
h_1^\mathcal{R} \odot h_2^\mathcal{R} ,   h_1^\mathcal{I} \odot  h_2^\mathcal{I}  ,\\
h_1^\mathcal{R}  \odot h_2^\mathcal{I} ,  h_1^\mathcal{I} \odot h_2^\mathcal{R}
]
\end{split}
\end{equation}

$ f_\mathbb{C^{\alpha}}$ can be seen as Distmult concatenated with the asymmetrical part of ComplEx and $ f_\mathbb{C^{\beta}}$ can be seen as RESCAL with unconstrained block diagonal relation matrices. 
%

\section{On the evaluation of relational models}
\label{sec:eval}

The SentEval framework \citep{Conneau2017} provides a general evaluation for transferable sentence representations, with open source evaluation code. One only needs to specify a sentence encoder function, and the framework performs classification tasks or relation prediction tasks using cross-validated logistic regression on embeddings or composed sentence embeddings. Tasks include sentiment analysis, entailment, textual similarity, textual relatedness, and paraphrase detection. These tasks are a rich way to train or evaluate sentence representations since in a triple $(s_1, R, s_2)$, we can see $(R,s_2)$ as a label for $s_1$ \citep{Baudi}. 
Unfortunately, the relational tasks hard-code the composition function from equation \ref{eq:vanillacomposition}.
From our previous analysis, we believe this composition function favors the use of contextual/lexical similarity rather than high-level reasoning and can penalize representations based on more semantic aspects.  This bias could harm research since semantic representation is an important next step for sentence embedding.  Training/evaluation datasets are also arguably flawed with respect to relational aspects since several recent studies \citep{Dasgupta2018,Poliak2018, Gururangan2018, Glockner2018} show that InferSent, despite being state of the art on SentEval evaluation tasks, has poor performance when dealing with asymmetrical tasks and non-additive composition of words.
%
%
In addition to providing new ways of training sentence encoders, we will also extend the SentEval evaluation framework with a more expressive composition function when dealing with relational transfer tasks, which improves results even when the sentence encoder was not trained with it.

\section{Experiments}
\label{sec:exp}
Our goal is to show that transferable sentence representation learning and relation prediction tasks can be improved  when our expressive compositions are used instead of the composition from equation~\ref{eq:vanillacomposition}.
We train our relational model adaptations on two relation prediction base tasks ($\mathcal T$), one supervised ($\mathcal{T}= \mathit{NLI}$)  and one unsupervised ($\mathcal{T}= \mathit{Disc}$) described below, and evaluate sentence/relation representations on base and transfer tasks using the SentEval framework in order to quantify the generalization capabilities of our models. 
Since we use minor modifications of InferSent and SentEval, our experiments are easily reproducible.

 \begin{table*}[t]
\centering
\small
\begin{tabular}{lllll}
\toprule
   name &     N &                         task &           C &               representation(s) used \\
\midrule
     MR &   11k &           sentiment (movies) &           2 &                        $h_1$ \\
   SUBJ &   10k &     subjectivity/objectivity &           2 &                        $h_1$ \\
   MPQA &   11k &             opinion polarity &           2 &                        $h_1$ \\
   TREC &    6k &                question-type &           6 &                        $h_1$ \\
 $\text{SICK}^{m}_s$ &   10k &                          NLI &           3 &                 $f_{m,s}(h_1,h_2)$ \\
   $\text{MRPC}^{m}_s$ &  4k &         paraphrase detection &           2 &  $(f_{m,s}(h_1,h_2)+(f_{m,s}(h_2,h_1))/2$ \\
   $\text{PDTB}^{m}_s$ &  17k &                  discursive relation &  5 &             $f_{m,s}(h_1,h_2)$ \\
 STS14 &  4.5k &                  similarity &  - &               $\text{cos}(h_1,h_2)$ \\
\bottomrule
\end{tabular}
 \caption{Transfer evaluation tasks. N = number of training examples; C = number of classes if applicable. $h_1, h_2$ are sentence representations, $f_{m,s}$ a composition function from section \ref{sec:trm}.}
\label{table:evaltasks}
 \end{table*}
\subsection{Training tasks}
\label{sec:tr_nli}
Natural language inference  ($\mathcal{T}$ = {\it NLI})'s goal is to predict whether the relation between two sentences (premise and hypothesis) is \textit{Entailment}, \textit{Contradiction} or \textit{Neutral}.
We use the combination of SNLI dataset \citep{Bowman} and MNLI dataset \citep{Williams2017}. We call {\it AllNLI} the resulting dataset of $1M$ examples.
\citet{Conneau2017} claim that NLI data allows universal sentence representation learning. They used the $f_{\odot,-}$ composition function with concatenated sentence representations in order to train their \textit{Infersent} model.

We also train on the prediction of discourse connectives between sentences/clauses  ($\mathcal{T}$ = {\it Disc}). Discourse connectives make discourse relations between sentences explicit. In the sentence \textit{I live in Paris but I'm often elsewhere}, the word \textit{but} highlights that there is a contrast between the two clauses it connects. 
We use Malmi et al.'s (\citeyear{Malmi2017}) dataset of selected $400k$ instances with $20$ discourse connectives (e.g. \textit{however}, \textit{for example}) with the provided train/dev/test split. This dataset has no other supervision than the list of 20 connectives.
\citet{Nie2017} used $f_{\odot,-}$ concatenated with the sum of sentence representations to train their model, \textit{DisSent}, on a similar task and showed that their encoder was general enough to perform well on Sent\-Eval tasks.  They use a dataset that is, at the time of writing, not publicly available.
%
\subsection{Evaluation tasks}
Table \ref{table:evaltasks} provides an overview of different transfer tasks that will be used for evaluation.
We added another relation prediction task, the PDTB coarse-grained implicit discourse relation task, to Sent\-Eval. This task involves predicting a discursive link between two sentences among $\{$Comparison, Contingency, Entity based coherence, Expansion, Temporal$\}$.
We followed the setup of \citet{Pitler2009}, without sampling negative examples in training.
MRPC, PDTB and SICK will be tested with two composition functions: besides SentEval composition $f_{\odot,-}$, we will use $f_{\mathcal{C}^{\beta},-}$ for transfer learning evaluation,
since it has the most general multiplicative interaction and it does not penalize models that do not learn a translation.
 For all tasks except STS14, a cross-validated logistic regression is used on the sentence or relation representation. 
The evaluation of the STS14 task relies on Pearson or Spearman correlation between cosine similarity and the target. We force the composition function to be symmetrical on the MRPC task since paraphrase detection should be invariant to permutation of input sentences.

\subsection{Setup}
We want to compare the different instances of $f$.
We follow the setup of Infersent \citep{Conneau2017}: we learn to encode sentences into $h$ with a bi-directional LSTM  using element-wise max pooling over time. 
The dimension size of $h$ is $4096$.  Word embeddings are fixed GloVe with 300 dimensions, trained on Common Crawl 840B.\footnote{https://nlp.stanford.edu/projects/glove/} Optimization is done with SGD and decreasing learning rate until convergence.

The only difference with regard to Infersent is the composition. Sentences are composed with six different compositions for training according to the following template:
\begin{equation}
f_{m,s,1,2}(h_1,h_2) = [f_m(h_1,h_2), f_s(h_1,h_2), h_1, h_2]
\end{equation}
$f_s$ (subtractive interaction) is in  $\{ f_{-}, f_{t} \}$,
$f_m$ (multiplicative interaction) is in $\{{f_\odot}, f_{\mathbb C ^{\alpha}}, f_{\mathbb C ^{\beta}} \}$. We do not consider $f_\mathbb{C}$ since it yielded inferior results in our early experiments using NLI and Sent\-Eval development sets.

$f_{m,s,1,2}(h_1,h_2) $ is fed directly to a softmax regression. Note that Infersent uses a multi-layer perceptron before the softmax, but uses only linear activations, so $f_{\odot,-,1,2}(h_1,h_2) $ is analytically equivalent to Infersent when $\mathcal{T}= \mathit{NLI}$ .

\subsection{Results}

\begin{table*}[htb]
\centering
\small
\begin{tabular}{lrrrrrrrrrr}
  \toprule
\multicolumn{10}{c}{Models trained on natural language inference ($\mathcal T=\mathit{NLI}$)}          \\ \midrule
  m,s         & MR            & SUBJ          & MPQA          & TREC          & $\text{MRPC}^{\odot}_{-}$           & $\text{PDTB}^{\odot}_{-}$          & $\text{SICK}^{\odot}_{-}$          & STS14         & $\mathcal T$  & AVG
  \\ \midrule
 $\odot, -$   &             81.2 &             92.7 &             90.4 &             89.6 &             76.1 &             46.7 &             86.6 &             69.5 &             84.2 &             79.1 \\
 $\alpha, -$  &  $\textbf{81.4}$ &  $\textbf{92.8}$ &             90.5 &             89.6 &             75.4 &             46.6 &             86.7 &             69.5 &             84.3 &             79.1 \\
 $\beta, -$   &             81.2 &             92.6 &             90.5 &             89.6 &               76 &             46.5 &             86.6 &             69.5 &             84.2 &             79.1 \\
 $\odot, t$   &             81.1 &             92.7 &             90.5 &  $\textbf{89.7}$ &  $\textbf{76.5}$ &             46.4 &             86.5 &  $\textbf{70.0}$ &  $\textbf{84.8}$ &  $\textbf{79.2}$ \\
 $\alpha, t$  &             81.3 &             92.6 &  $\textbf{90.6}$ &             89.2 &             76.2 &             47.2 &             86.5 &  $\textbf{70.0}$ &             84.6 &  $\textbf{79.2}$ \\
 $\beta, t$   &             81.2 &             92.7 &             90.4 &             88.5 &             75.8 &  $\textbf{47.3}$ &  $\textbf{86.8}$ &             69.8 &             84.2 &             79.1 \\

  \bottomrule
\end{tabular}

\caption{SentEval and base task evaluation results for the models
  trained on natural language inference ($\mathcal T=\mathit{NLI}$);
  AllNLI is used for training. All scores are accuracy percentages,
  except STS14, which is Pearson correlation percentage. AVG denotes the average of the SentEval scores.} 
\label{table:resnli}
\end{table*}



 \begin{table*}[htb]
\centering
\small
\begin{tabular}{lrrrrrrrrrr}
\toprule
\multicolumn{10}{c}{Models trained on discourse connective prediction ($\mathcal T=\mathit{Disc}$)}          \\ \midrule
  m,s         & MR            & SUBJ          & MPQA          & TREC          & $\text{MRPC}^{\odot}_{-}$           & $\text{PDTB}^{\odot}_{-}$          & $\text{SICK}^{\odot}_{-}$          & STS14         & $\mathcal T$ & AVG\\ \midrule
 $\odot, -$   &  $\textbf{80.4}$ &             92.7 &             90.2 &             89.5 &             74.5 &             47.3 &             83.2 &             57.9 &             35.7 &               77 \\
 $\alpha, -$  &  $\textbf{80.4}$ &  $\textbf{92.9}$ &             90.2 &             90.2 &               75 &             47.9 &             83.3 &             57.8 &             35.9 &             77.2 \\
 $\beta, -$   &             80.2 &             92.8 &             90.2 &             88.4 &             74.9 &             47.5 &             82.9 &             57.7 &             35.9 &             76.8 \\
 $\odot, t$   &             80.2 &             92.8 &             90.2 &  $\textbf{90.4}$ &             74.6 &  $\textbf{48.5}$ &             83.4 &  $\textbf{58.6}$ &  $\textbf{36.1}$ &  $\textbf{77.3}$ \\
 $\alpha, t$  &             80.2 &  $\textbf{92.9}$ &  $\textbf{90.3}$ &             90.3 &  $\textbf{75.1}$ &             47.8 &             83.2 &             58.3 &  $\textbf{36.1}$ &  $\textbf{77.3}$ \\
 $\beta, t$   &             80.2 &             92.8 &  $\textbf{90.3}$ &             89.7 &             74.4 &             47.9 &  $\textbf{83.7}$ &             58.2 &             35.7 &             77.2 \\ \bottomrule
\end{tabular}
\caption{SentEval and base task evaluation results for the models
  trained on discourse connective prediction
  ($\mathcal T = \mathit{Disc}$). All scores are accuracy percentages,
  except STS14, which is Pearson correlation percentage.  AVG denotes the average of the SentEval scores.  }
\label{table:resdisc}
\end{table*}

\begin{table*}[htb]
\centering
\small
\begin{tabular}{lrrrrrrrrrrr}
  \toprule
\multicolumn{9}{c}{Comparison models}          \\ \midrule
model           & MR            & SUBJ          & MPQA          & TREC          & $\text{MRPC}^{\odot}_{-}$           & $\text{PDTB}^{\odot}_{-}$          & $\text{SICK}^{\odot}_{-}$          & STS14 &AVG      \\ \midrule
Infersent&81.1&92.4&90.2&88.2&76.2&46.7-&86.3&70&78.9\\
SkipT  &76.5&93.6&87.1&92.2&73& -          &82.3&29&- \\
  BoW  &77.2&91.2&87.9&83&72.2&43.9&78.4&54.6&73.6 \\ \bottomrule
\end{tabular}
\caption{Comparison models from previous work. InferSent represents the original results from \citet{Conneau2017}, SkipT is SkipThought from \citet{Kiros2015}, and BoW is our re-evaluation of GloVe Bag of Words from \citet{Conneau2017}. AVG denotes the average of the SentEval scores..
  }
\label{table:comparison}
\end{table*}
\begin{table*}[t]
\centering
\small
\begin{tabular}{lrrrrrrrr}
  \toprule
& \multicolumn{4}{c}{$\mathcal T=\mathit{Disc}$} & \multicolumn{4}{c}{$\mathcal T=\mathit{NLI}$} \\ \cmidrule(lr{0.2em}){2-5} \cmidrule(lr{0.2em}){6-9}
m,s         &  $\text{MRPC}^{\beta}_{-}$         &  $\text{PDTB}^{\beta}_{-}$         &  $\text{SICK}^{\beta}_{-}$   &AVG      &  $\text{MRPC}^{\beta}_{-}$         &  $\text{PDTB}^{\beta}_{-}$         &  $\text{SICK}^{\beta}_{-}$     &AVG   \\ \midrule
 $\odot, -$   &             74.8 &             48.2 &             83.6 &             68.9 &  $\textbf{76.2}$ &             47.2 &             86.9 &             70.1 \\
 $\alpha, -$  &             74.9 &  $\textbf{49.3}$ &             83.8 &  $\textbf{69.3}$ &             75.9 &             47.1 &             86.9 &               70 \\
 $\beta, -$   &               75 &             48.8 &             83.4 &             69.1 &             75.8 &               47 &               87 &             69.9 \\
 $\odot, t$   &             74.9 &             48.7 &             83.6 &             69.1 &  $\textbf{76.2}$ &  $\textbf{47.8}$ &             86.8 &             70.3 \\
 $\alpha, t$  &  $\textbf{75.2}$ &             48.6 &             83.5 &             69.1 &  $\textbf{76.2}$ &             47.6 &  $\textbf{87.3}$ &  $\textbf{70.4}$ \\
 $\beta, t$   &             74.6 &             48.9 &  $\textbf{83.9}$ &             69.1 &  $\textbf{76.2}$ &  $\textbf{47.8}$ &               87 &             70.3 \\
  \bottomrule
\end{tabular}
\caption{Results for sentence relation tasks using an alternative
  composition function ($f_{\mathbb{C}^{\beta}, -}$) during 
  evaluation. AVG denotes the average of the three tasks.}
\label{table:alteval}
\end{table*}

Having run several experiments with different initializations, the
standard deviations between them do not seem to be negligible. We
decided to take these into account when reporting scores, contrary to
previous work \citep{Kiros2015,Conneau2017}: we average the scores of
6 distinct runs for each task and use standard deviations under
normality assumption to compute significance. Table \ref{table:resnli}
shows model scores for $\mathcal{T}= \mathit{NLI}$, while Table
\ref{table:resdisc} shows scores for $\mathcal T =Disc $. For comparison, Table
\ref{table:comparison} shows a number of important models from
previous work. Finally, in Table \ref{table:alteval}, we present results for sentence relation tasks that use an alternative composition function ($f_{\mathbb{C}^{\beta}, -}$) instead of the standard composition function used in SentEval.




For sentence representation learning, the baseline, $f_\odot- $  composition already performs rather well, being on par with the InferSent scores of the original paper, as would be expected. However, macro-averaging all accuracies, it is the second worst performing model. 
$f_{\mathbb{C}^\alpha,t,1,2}$ is the best performing model, and all three best models use the translation ($s=t$).
On relational transfer tasks, training with $f_{\mathbb{C}^\alpha,t,1,2}$ and using complex $\mathbb{C}^\beta$ for transfer (Table \ref{table:alteval})  always outperforms the baseline ($f_{ \odot,-,1,2}$ with $\odot -$ composition in Tables \ref{table:resnli} and \ref{table:resdisc}). Averaging accuracies of those transfer tasks, this result is significant for both training tasks at level $p<0.05$ (using Bonferroni correction accounting for the 5 comparisons). 
On base tasks and the average of non-relational transfer tasks (MR, MPQA, SUBJ, TREC),  our proposed compositions are on average slightly better than  $f_{ \odot,-,1,2}$.
Representations learned with our proposed compositions can still be compared with  simple cosine similarity: all three methods using the translational composition ($s=t$)  very significantly outperform the baseline (significant at level $p<0.01$ with Bonferroni correction) on STS14 for $\mathcal T= \mathit{NLI}$. 
Thus, we  believe $f_{\mathcal{C}^\alpha,t,1,2}$ 
has more robust results and could be a better default choice than $f_{ \odot,-,1,2}$ as composition for representation learning.
\footnote{Note that our compositions are also beneficial with regard to convergence speed:  on average, each of our proposed compositions needed less epochs to converge than the baseline $f_{ \odot,-,1,2}$, for both training tasks.}

Additionally, using $\mathbb{C}^\beta$ (Table \ref{table:alteval}) instead of $\odot$ (Tables \ref{table:resnli}  and \ref{table:resdisc}) for transfer learning
in relational transfer tasks (PDTB, MRPC, SICK) yields a significant improvement on average, even when $m=\odot$ was used for training ($p<0.001$). Therefore, we believe $f_{\mathbb{C}^\beta,-}$ is an interesting composition for inference or evaluation of models regardless of how they were trained.
%
 
\section{Related work}
\label{sec:rl}
There are numerous interactions between SRL and NLP. 
We believe that our framework merges two specific lines of work: relation prediction and modeling textual relational tasks.


Some previous NLP work focused on composition functions for relation prediction between text fragments, even though they ignored SRL and only dealt with word units. Word2vec \citep{Mikolov2013} has sparked a great interest for this task with word analogies in the latent space. 
%
%
\citet{Levy2014} explored different scoring functions between words, notably for analogies. Hypernymy relations were also studied, by \citet{Chang2017} and \citet{Fu2014}. \citet{Levy2015a} proposed tailored scoring functions. Even the skipgram model \citep{Mikolov2013} can be formulated as finding relations between context and target words. We did not  empirically explore textual relational learning at the word level, but we believe that it would fit in our framework, and could be tested in future studies.
 Numerous approaches \citep{conf/acl/ChenZLWJI17,Seok2016,gong2018natural,pair2vec} were proposed to predict inference relations between sentences, but don't explicitely use sentence embeddings. Instead, they encode sentences jointly, possibly with the help of previously cited word compositions, therefore it would also be interesting to try applying our techniques within their framework.


Some modeling aspects of textual relational learning have been formally investigated by \citet{Baudi}. They noticed the genericity of relational problems and explored multi-task and transfer learning on relational tasks. Their work is complementary to ours since  their framework unifies tasks while ours unifies composition functions.  Subsequent approaches use relational tasks for training and evaluation on specific datasets \citep{Conneau2017,Nie2017}.

\section{Conclusion}
We  have demonstrated that a number of existing models used for textual relational learning rely on composition functions that are already used in Statistical Relational Learning. By taking into account previous insights from SRL, we proposed new composition functions and evaluated them.
These composition functions are all simple to implement and we hope that it will become standard to try them on relational problems. Larger scale data might leverage these more expressive compositions, as well as more compositional, asymmetric, and arguably more realistic datasets \citep{Dasgupta2018,Gururangan2018}.
Finally, our compositions can also be helpful to improve interpretability of embeddings, since they can help measure relation prediction asymmetry. Analogies through translations helped interpreting word embeddings, and perhaps anlyzing our learned $t$ translation could help interpreting sentence embeddings.

\newpage

\bibliography{naacl2018}

\begin{thebibliography}{42}
\providecommand{\natexlab}[1]{#1}
\providecommand{\url}[1]{\texttt{#1}}
\expandafter\ifx\csname urlstyle\endcsname\relax
  \providecommand{\doi}[1]{doi: #1}\else
  \providecommand{\doi}{doi: \begingroup \urlstyle{rm}\Url}\fi

\bibitem[Ampomah et~al.(2016)Ampomah, Park, and Lee]{Ampomah2016}
Isaac K~E Ampomah, Seong-bae Park, and Sang-jo Lee.
\newblock {A Sentence-to-Sentence Relation Network for Recognizing Textual
  Entailment}.
\newblock \emph{World Academy of Science, Engineering and Technology
  International Journal of Computer and Information Engineering}, 10\penalty0
  (12):\penalty0 1955--1958, 2016.

\bibitem[Baudi{\v{s}} et~al.(2016)Baudi{\v{s}}, Pichl, Vysko{\v{c}}il, and
  {\v{S}}ediv{\'{y}}]{Baudi}
Petr Baudi{\v{s}}, Jan Pichl, Tom{\'{a}}{\v{s}} Vysko{\v{c}}il, and Jan
  {\v{S}}ediv{\'{y}}.
\newblock {Sentence Pair Scoring: Towards Unified Framework for Text
  Comprehension}.
\newblock 2016.
\newblock URL \url{http://arxiv.org/abs/1603.06127}.

\bibitem[Bordes et~al.(2013{\natexlab{a}})Bordes, Glorot, Weston, and
  Bengio]{Bordes2013a}
Antoine Bordes, Xavier Glorot, Jason Weston, and Yoshua Bengio.
\newblock {A Semantic Matching Energy Function for Learning with
  Multi-relational Data}.
\newblock \emph{Machine Learning}, 2013{\natexlab{a}}.
\newblock ISSN 0885-6125.
\newblock \doi{10.1007/s10994-013-5363-6}.
\newblock URL \url{http://arxiv.org/abs/1301.3485}.

\bibitem[Bordes et~al.(2013{\natexlab{b}})Bordes, Usunier, Weston, and
  Yakhnenko]{Bordes2013}
Antoine Bordes, Nicolas Usunier, Jason Weston, and Oksana Yakhnenko.
\newblock {Translating Embeddings for Modeling Multi-Relational Data}.
\newblock \emph{Advances in NIPS}, 26:\penalty0 2787--2795, 2013{\natexlab{b}}.
\newblock ISSN 10495258.
\newblock \doi{10.1007/s13398-014-0173-7.2}.

\bibitem[Bowman et~al.(2015)Bowman, Angeli, Potts, and Manning]{Bowman}
Samuel~R Bowman, Gabor Angeli, Christopher Potts, and Christopher~D Manning.
\newblock {A large annotated corpus for learning natural language inference}.
\newblock \emph{Proceedings of the 2015 Conference on Empirical Methods in
  Natural Language Processing,Lisbon, Portugal, 17-21 September 2015},
  \penalty0 (September):\penalty0 632--642, 2015.
\newblock ISSN 9781941643327.

\bibitem[Chang et~al.(2017)Chang, Wang, Vilnis, and McCallum]{Chang2017}
Haw-Shiuan Chang, ZiYun Wang, Luke Vilnis, and Andrew McCallum.
\newblock {Unsupervised Hypernym Detection by Distributional Inclusion Vector
  Embedding}.
\newblock 2017.
\newblock URL \url{http://arxiv.org/abs/1710.00880}.

\bibitem[Chen et~al.(2017{\natexlab{a}})Chen, Peterson, and
  Griffiths]{Chen2017a}
Dawn Chen, Joshua~C. Peterson, and Thomas~L. Griffiths.
\newblock Evaluating vector-space models of analogy.
\newblock \emph{CoRR}, abs/1705.04416, 2017{\natexlab{a}}.

\bibitem[Chen et~al.(2017{\natexlab{b}})Chen, Zhu, Ling, Wei, Jiang, and
  Inkpen]{conf/acl/ChenZLWJI17}
Qian Chen, Xiaodan Zhu, Zhen-Hua Ling, Si~Wei, Hui Jiang, and Diana Inkpen.
\newblock Enhanced lstm for natural language inference.
\newblock In Regina Barzilay and Min-Yen Kan (eds.), \emph{ACL (1)}, pp.\
  1657--1668. Association for Computational Linguistics, 2017{\natexlab{b}}.
\newblock ISBN 978-1-945626-75-3.
\newblock URL
  \url{http://dblp.uni-trier.de/db/conf/acl/acl2017-1.html#ChenZLWJI17}.

\bibitem[Conneau et~al.(2017)Conneau, Kiela, Schwenk, Barrault, and
  Bordes]{Conneau2017}
Alexis Conneau, Douwe Kiela, Holger Schwenk, Loic Barrault, and Antoine Bordes.
\newblock {Supervised Learning of Universal Sentence Representations from
  Natural Language Inference Data}.
\newblock \emph{Emnlp}, 2017.

\bibitem[Dasgupta et~al.(2018)Dasgupta, Guo, Stuhlm{\"{u}}ller, Gershman, and
  Goodman]{Dasgupta2018}
Ishita Dasgupta, Demi Guo, Andreas Stuhlm{\"{u}}ller, Samuel~J. Gershman, and
  Noah~D. Goodman.
\newblock {Evaluating Compositionality in Sentence Embeddings}.
\newblock \penalty0 (2011), 2018.
\newblock URL \url{http://arxiv.org/abs/1802.04302}.

\bibitem[Devlin et~al.(2018)Devlin, Chang, Lee, and Toutanova]{devlin2018bert}
Jacob Devlin, Ming-Wei Chang, Kenton Lee, and Kristina Toutanova.
\newblock Bert: Pre-training of deep bidirectional transformers for language
  understanding.
\newblock \emph{arXiv preprint arXiv:1810.04805}, 2018.

\bibitem[Fu et~al.(2014)Fu, Guo, Qin, Che, Wang, and Liu]{Fu2014}
Ruiji Fu, Jiang Guo, Bing Qin, Wanxiang Che, Haifeng Wang, and Ting Liu.
\newblock {Learning Semantic Hierarchies via Word Embeddings}.
\newblock \emph{Acl}, pp.\  1199--1209, 2014.

\bibitem[Getoor \& Taskar(2007)Getoor and Taskar]{Getoor:2007:ISR:1296231}
Lise Getoor and Ben Taskar.
\newblock \emph{{Introduction to Statistical Relational Learning (Adaptive
  Computation and Machine Learning)}}.
\newblock The MIT Press, 2007.
\newblock ISBN 0262072882.

\bibitem[Glockner et~al.(2018)Glockner, Shwartz, and Goldberg]{Glockner2018}
Max Glockner, Vered Shwartz, and Yoav Goldberg.
\newblock {Breaking NLI Systems with Sentences that Require Simple Lexical
  Inferences}.
\newblock \emph{Proceedings of the 56th Annual Meeting of the Association for
  Computational Linguistics (Short Papers)}, \penalty0 (3):\penalty0 1--6,
  2018.
\newblock URL \url{http://arxiv.org/abs/1805.02266}.

\bibitem[Gong et~al.(2018)Gong, Luo, and Zhang]{gong2018natural}
Yichen Gong, Heng Luo, and Jian Zhang.
\newblock Natural language inference over interaction space.
\newblock In \emph{International Conference on Learning Representations}, 2018.
\newblock URL \url{https://openreview.net/forum?id=r1dHXnH6-}.

\bibitem[Gururangan et~al.(2018)Gururangan, Swayamdipta, Levy, Schwartz,
  Bowman, and Smith]{Gururangan2018}
Suchin Gururangan, Swabha Swayamdipta, Omer Levy, Roy Schwartz, Samuel~R.
  Bowman, and Noah~A. Smith.
\newblock {Annotation Artifacts in Natural Language Inference Data}.
\newblock 2018.
\newblock URL \url{http://arxiv.org/abs/1803.02324}.

\bibitem[Hooda \& Kosseim(2017)Hooda and Kosseim]{Hooda2017}
Sohail Hooda and Leila Kosseim.
\newblock {Argument Labeling of Explicit Discourse Relations using LSTM Neural
  Networks}.
\newblock 2017.
\newblock URL \url{http://arxiv.org/abs/1708.03425}.

\bibitem[Jernite et~al.(2017)Jernite, Bowman, and Sontag]{Jernite2017}
Yacine Jernite, Samuel~R. Bowman, and David Sontag.
\newblock {Discourse-Based Objectives for Fast Unsupervised Sentence
  Representation Learning}.
\newblock 2017.
\newblock URL \url{http://arxiv.org/abs/1705.00557}.

\bibitem[Joshi et~al.(2018)Joshi, Choi, Levy, Weld, and Zettlemoyer]{pair2vec}
Mandar Joshi, Eunsol Choi, Omer Levy, Daniel~S. Weld, and Luke Zettlemoyer.
\newblock pair2vec: Compositional word-pair embeddings for cross-sentence
  inference.
\newblock \emph{CoRR}, abs/1810.08854, 2018.
\newblock URL \url{http://arxiv.org/abs/1810.08854}.

\bibitem[Kiros et~al.(2015)Kiros, Zhu, Salakhutdinov, Zemel, Urtasun, Torralba,
  and Fidler]{Kiros2015}
Ryan Kiros, Yukun Zhu, Ruslan~R Salakhutdinov, Richard Zemel, Raquel Urtasun,
  Antonio Torralba, and Sanja Fidler.
\newblock {Skip-thought vectors}.
\newblock In \emph{Advances in neural information processing systems}, pp.\
  3294--3302, 2015.

\bibitem[Lacroix et~al.(2018)Lacroix, Usunier, and
  Obozinski]{Lacroix2018CanonicalTD}
Timoth{\'e}e Lacroix, Nicolas Usunier, and Guillaume Obozinski.
\newblock Canonical tensor decomposition for knowledge base completion.
\newblock In \emph{ICML}, 2018.

\bibitem[Levy \& Goldberg(2014)Levy and Goldberg]{Levy2014}
Omer Levy and Yoav Goldberg.
\newblock {Linguistic Regularities in Sparse and Explicit Word
  Representations}.
\newblock \emph{Proceedings of the Eighteenth Conference on Computational
  Natural Language Learning}, pp.\  171--180, 2014.
\newblock \doi{10.3115/v1/W14-1618}.
\newblock URL \url{http://aclweb.org/anthology/W14-1618}.

\bibitem[Levy et~al.(2015)Levy, Remus, Biemann, and Dagan]{Levy2015a}
Omer Levy, Steffen Remus, Chris Biemann, and Ido Dagan.
\newblock {Do Supervised Distributional Methods Really Learn Lexical Inference
  Relations?}
\newblock \emph{Naacl-2015}, pp.\  970--976, 2015.
\newblock URL \url{http://www.aclweb.org/anthology/N/N15/N15-1098.pdf}.

\bibitem[Liu et~al.(2017)Liu, Wu, and Yang]{Liu2017a}
Hanxiao Liu, Yuexin Wu, and Yiming Yang.
\newblock {Analogical Inference for Multi-Relational Embeddings}.
\newblock \emph{Icml}, 2017.
\newblock ISSN 1938-7228.
\newblock URL \url{http://arxiv.org/abs/1705.02426}.

\bibitem[Logeswaran \& Lee(2018)Logeswaran and Lee]{Logeswaran2018}
Lajanugen Logeswaran and Honglak Lee.
\newblock {An efficient framework for learning sentence representations}.
\newblock pp.\  1--16, 2018.
\newblock URL \url{http://arxiv.org/abs/1803.02893}.

\bibitem[Logeswaran et~al.(2016)Logeswaran, Lee, and Radev]{Logeswaran2016}
Lajanugen Logeswaran, Honglak Lee, and Dragomir Radev.
\newblock {Sentence Ordering using Recurrent Neural Networks}.
\newblock pp.\  1--15, 2016.
\newblock URL \url{http://arxiv.org/abs/1611.02654}.

\bibitem[Malmi et~al.(2017)Malmi, Pighin, Krause, and Kozhevnikov]{Malmi2017}
Eric Malmi, Daniele Pighin, Sebastian Krause, and Mikhail Kozhevnikov.
\newblock {Automatic Prediction of Discourse Connectives}.
\newblock 2017.
\newblock URL \url{http://arxiv.org/abs/1702.00992}.

\bibitem[Mikolov et~al.(2013)Mikolov, Chen, Corrado, and Dean]{Mikolov2013}
Tomas Mikolov, Kai Chen, Greg Corrado, and Jeffrey Dean.
\newblock {Distributed Representations of Words and Phrases and their
  Compositionality}.
\newblock \emph{Nips}, pp.\  1--9, 2013.
\newblock ISSN 10495258.
\newblock \doi{10.1162/jmlr.2003.3.4-5.951}.

\bibitem[Mou et~al.(2015)Mou, Men, Li, Xu, Zhang, Yan, and Jin]{Mou2015}
Lili Mou, Rui Men, Ge~Li, Yan Xu, Lu~Zhang, Rui Yan, and Zhi Jin.
\newblock {Natural Language Inference by Tree-Based Convolution and Heuristic
  Matching}.
\newblock pp.\  130--136, 2015.
\newblock URL \url{http://arxiv.org/abs/1512.08422}.

\bibitem[Mueller \& Thyagarajan(2016)Mueller and
  Thyagarajan]{mueller2016siamese}
Jonas Mueller and Aditya Thyagarajan.
\newblock {Siamese Recurrent Architectures for Learning Sentence Similarity.}
\newblock In \emph{AAAI}, pp.\  2786--2792, 2016.

\bibitem[Nickel et~al.(2011)Nickel, Tresp, and Kriegel]{Nickel2011}
Maximilian Nickel, Volker Tresp, and Hans-Peter Kriegel.
\newblock {A Three-Way Model for Collective Learning on Multi-Relational Data}.
\newblock \emph{Icml}, pp.\  809----816, 2011.

\bibitem[Nie et~al.(2017)Nie, Bennett, and Goodman]{Nie2017}
Allen Nie, Erin~D. Bennett, and Noah~D. Goodman.
\newblock {DisSent: Sentence Representation Learning from Explicit Discourse
  Relations}.
\newblock 2017.
\newblock URL \url{http://arxiv.org/abs/1710.04334}.

\bibitem[Pitler et~al.(2009)Pitler, Louis, and Nenkova]{Pitler2009}
Emily Pitler, Annie Louis, and Ani Nenkova.
\newblock {Automatic sense prediction for implicit discourse relations in
  text}.
\newblock \emph{Proceedings of the Joint Conference of the 47th Annual Meeting
  of the ACL and the 4th International Joint Conference on Natural Language
  Processing of the AFNLP Volume 2 ACLIJCNLP 09}, 2\penalty0 (August):\penalty0
  683--691, 2009.
\newblock \doi{10.3115/1690219.1690241}.
\newblock URL \url{http://www.aclweb.org/anthology/P/P09/P09-1077}.

\bibitem[Poliak et~al.(2018)Poliak, Naradowsky, Haldar, Rudinger, and
  Durme]{Poliak2018}
Adam Poliak, Jason Naradowsky, Aparajita Haldar, Rachel Rudinger, and
  Benjamin~Van Durme.
\newblock {Hypothesis Only Baselines in Natural Language Inference}.
\newblock \emph{Proceedings of the 7th Joint Conference on Lexical and
  Computational Semantics}, \penalty0 (1):\penalty0 180--191, 2018.

\bibitem[Radford(2018)]{Radford2018ImprovingLU}
Alec Radford.
\newblock Improving language understanding by generative pre-training.
\newblock 2018.

\bibitem[Seo et~al.(2016)Seo, Kembhavi, Farhadi, and
  Hajishirzi]{DBLP:journals/corr/SeoKFH16}
Min~Joon Seo, Aniruddha Kembhavi, Ali Farhadi, and Hannaneh Hajishirzi.
\newblock Bidirectional attention flow for machine comprehension.
\newblock \emph{CoRR}, abs/1611.01603, 2016.
\newblock URL \url{http://arxiv.org/abs/1611.01603}.

\bibitem[Seok et~al.(2016)Seok, Song, Park, Kim, and Kim]{Seok2016}
Miran Seok, Hye-Jeong Song, Chan-Young Park, Jong-Dae Kim, and Yu-Seop Kim.
\newblock {Named Entity Recognition using Word Embedding as a Feature 1}.
\newblock \emph{International Journal of Software Engineering and Its
  Applications}, 10\penalty0 (2):\penalty0 93--104, 2016.
\newblock ISSN 1738-9984.
\newblock \doi{10.14257/ijseia.2016.10.2.08}.
\newblock URL \url{http://dx.doi.org/10.14257/ijseia.2016.10.2.08}.

\bibitem[Socher et~al.(2013)Socher, Chen, Manning, Chen, and Ng]{Socher2013}
Richard Socher, Danqi Chen, Christopher Manning, Danqi Chen, and Andrew Ng.
\newblock {Reasoning With Neural Tensor Networks for Knowledge Base
  Completion}.
\newblock In \emph{Neural Information Processing Systems (2003)}, pp.\
  926--934, 2013.
\newblock URL
  \url{https://nlp.stanford.edu/pubs/SocherChenManningNg{\_}NIPS2013.pdf}.

\bibitem[Tai et~al.(2015)Tai, Socher, and Manning]{Tai2015}
Kai~Sheng Tai, Richard Socher, and Christopher~D. Manning.
\newblock {Improved Semantic Representations From Tree-Structured Long
  Short-Term Memory Networks}.
\newblock 2015.
\newblock ISSN 9781941643723.
\newblock \doi{10.1515/popets-2015-0023}.
\newblock URL \url{http://arxiv.org/abs/1503.00075}.

\bibitem[Trouillon et~al.(2016)Trouillon, Welbl, Riedel, Gaussier, and
  Bouchard]{Trouillon2016}
Th{\'{e}}o Trouillon, Johannes Welbl, Sebastian Riedel, {\'{E}}ric Gaussier,
  and Guillaume Bouchard.
\newblock {Complex Embeddings for Simple Link Prediction}.
\newblock In \emph{Proceedings of the 33nd International Conference on Machine
  Learning}, volume~48, 2016.
\newblock ISBN 9781510829008.
\newblock URL \url{http://arxiv.org/abs/1606.06357}.

\bibitem[Williams et~al.(2017)Williams, Nangia, and Bowman]{Williams2017}
Adina Williams, Nikita Nangia, and Samuel~R. Bowman.
\newblock {A Broad-Coverage Challenge Corpus for Sentence Understanding through
  Inference}.
\newblock 2017.
\newblock URL \url{http://arxiv.org/abs/1704.05426}.

\bibitem[Yang et~al.(2015)Yang, Lee, Park, and Rim]{Yang2015}
Min~Chul Yang, Do~Gil Lee, So~Young Park, and Hae~Chang Rim.
\newblock {Knowledge-based question answering using the semantic embedding
  space}.
\newblock \emph{Expert Systems with Applications}, 42\penalty0 (23):\penalty0
  9086--9104, 2015.
\newblock ISSN 09574174.
\newblock \doi{10.1016/j.eswa.2015.07.009}.
\newblock URL \url{http://dx.doi.org/10.1016/j.eswa.2015.07.009}.

\end{thebibliography}
\bibliographystyle{iclr2019_conference} 

\end{document}